\documentclass[journal,onecolumn]{IEEEtran}

  \usepackage{graphicx}
  \usepackage[export]{adjustbox}
  \usepackage[style=ieee]{biblatex}
  \usepackage{amssymb, amsmath}
   
\begin{document}

  \title{Neural Twins Talk \& Alternative Calculations}

    \author{\IEEEauthorblockN{Zanyar Zohourianshahzadi \& Jugal K. Kalita}
    \par
    \IEEEauthorblockA{
    Department of Computer Science\\
    University of Colorado Colorado Springs\\
    1420 Austin Bluffs Pkwy\\ Colorado Springs, Colorado 80918\\
    Email: \{zzohouri,jkalita\}@uccs.edu}
    }

  \maketitle

  \begin{abstract}
  Inspired by how the human brain employs more neural pathways when increasing the focus on a subject, 
  we introduce a novel twin cascaded attention model
  that outperforms a state-of-the-art image captioning model that was originally implemented using one channel of attention for the visual grounding task. 
  Visual grounding ensures the existence of words in the caption sentence that are grounded into a particular region in the input image.
  After a deep learning model is trained on visual grounding task, the model employs the learned patterns regarding
  the visual grounding and the order of objects in the caption sentences, when generating captions.
  We report the results of our experiments in three image captioning tasks on the COCO dataset.
  The results are reported using standard image captioning metrics to show the improvements achieved by our model over the previous image captioning model.
  The results gathered from our experiments suggest that employing more parallel attention pathways in a deep neural network leads to higher performance.
  Our implementation of NTT is publicly available at: https://github.com/zanyarz/NeuralTwinsTalk.
  \end{abstract}

  \IEEEpeerreviewmaketitle

  \section{Introduction}\label{sec1}
\par
Inspired by how the human brain employs a higher number of neural pathways when describing a highly focused subject,
we show that deep attentive models used for the main vision-language task of image captioning, could be extended to achieve better performance.
Image captioning bridges a gap between computer vision and natural language processing. 
Automated image captioning is used as a tool
to eliminate the need for human agent for creating descriptive captions for unseen images.
Automated image captioning is challenging and yet interesting.
One reason is that AI based systems capable of
generating sentences that describe an input image could
be used in a wide variety of tasks beyond generating captions for unseen images
found on web or uploaded to social media. For example, in biology and medical sciences, these
systems could provide researchers and physicians with a brief linguistic
description of relevant images, potentially expediting their work.
\par
\begin{figure}[ht!]
    \centering
    \includegraphics[width=10cm, height=7cm,frame={\fboxrule} {-\fboxrule}]{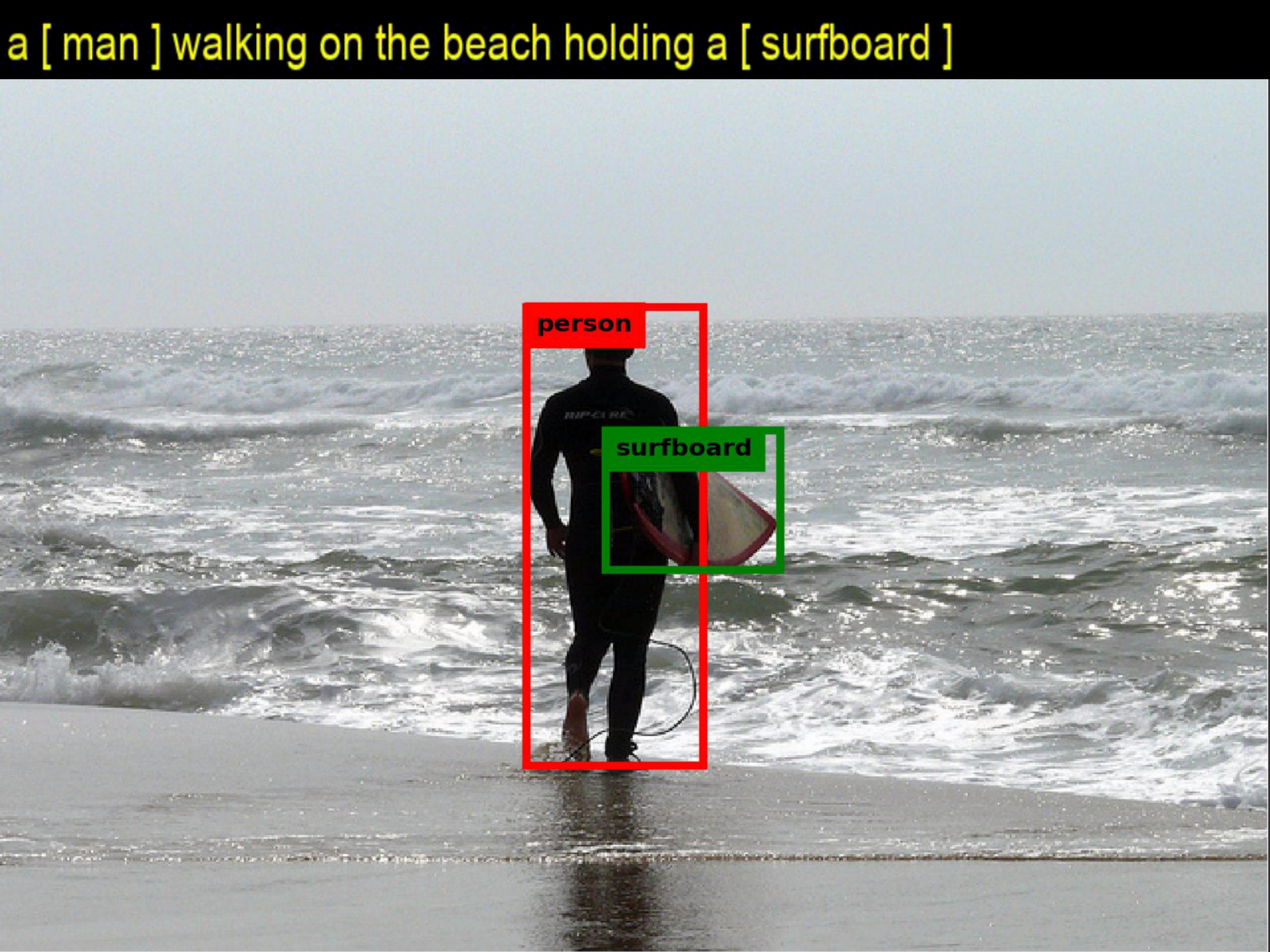}
    \caption{Example of generated caption for input image.
    Caption was generated by Neural Twins Talk, described in this paper in novel object
    detection task. The words that are placed inside brackets
    are the words that are visually grounded into a particular region in the image.}
    \label{fig1}
\end{figure}
\par
In this work, we improve the previous implementation of a state-of-the-art
image captioning model called Neural Baby Talk \cite{Lu2018NBT}, that ensures the ``visual grounding'' of the
generated words in the caption.
Neural Baby Talk is the first deep learning model to generate captions
containing words that relate to specific regions in an input image.
Fig. \ref{fig1} explains the visual grounding task.
A caption is generated with visual words detected and shown between brackets.
Intuitively, it makes sense to ensure the visual grounding of the words in the caption
that is being generated by the model. This is because
even humans tend to use visual representations
of different parts of an image to describe the whole content of an image.
In this paper, we show that our twin cascaded
attention model, which employs two parallel attention channels, improves the quality of generated captions in comparison with a similar model with one attention channel.
We refer to our method as Neural Twins Talk (NTT). 
Our contributions in this work can be summarized as the following.
\par
\begin{itemize}
    \item We show that deep learning models deploying attention mechanism \cite{ORG_Attention}
    on long short-term memory networks (LSTM) \cite{LSTM} could be improved using a novel
    twin cascaded attention method that we explain in detail in Section \ref{3.3}. We show this by
    improving NBT \cite{Lu2018NBT} and Bottom-up and Top-down Attention model \cite{Anderson2017up-down}.
    \item We introduce cascaded adaptive gates. In our model, we use these adaptive gates to improve the performance
    of the language models in the twin cascaded attention model. The values of adaptive gates
    are added to each other right before they are applied to the context of each language LSTM.
    This mechanism ensures that the next language LSTM in our model becomes aware of the attention in the
    previous language LSTMs.
    \item We show that by increasing the dropout rate \cite{Hinton2014_dropout}
    for the second and third language LSTMs in the proposed twin cascaded
    attention model, we avoid overfitting successfully and we create a
    refinement effect over the generated captions by creating a meta hypothesis vector.
    We explain this meta hypothesis and how it is calculated in Section \ref{3.3}.
    \item We improve the visual sentinel that was previously used in NBT. We achieve this by performing
    a non-linear transformation over the context vector coming from the last language LSTM in our model.
    This is similar to how it is calculated in NBT, except that we use the context
    vector from the joint LSTM rather than from the first one in our model.
    \item Instead of modifying the architectural change process while implementing the Neural Twins Talk algorithm, which utilizes parallel neural attention channels,
    we use an alternative calculation method.
    We show that the twin cascaded attention model could be utilized over
    various architectures, only via employing alternative calculation methods. This is further discussed in Section \ref{3.4}.
\end{itemize}
\par
The results of our experiments show that a deep model with twin cascaded attention performs better than a deep model
with a single channel of attention. At the same time, the results of our experiments
show that the twin cascaded attention model performs better than attention models
with a single channel of attention in bigger datasets where we have more training data available.
\par
We also perform experiments in MS-COCO with Bottom-up and Top-down Attention model \cite{Anderson2017up-down},
and we show that by employing alternative calculation methods over the same
parallel attention channel expansion technique, we could achieve better performance.
\par
Neural architecture search (NAS) techniques have been employed for
finding an optimal architecture in a search space of candidate architectures.
A good survey of NAS techniques was performed by Elsken et al. \cite{NAS_Survey_JMLR}. Instead of performing a search in a search space
including various architectures, we offer a new search space that includes various calculation methods over the same architecture.
\par
This could indeed lead to a broader search space for NAS techniques. Considering the fact that the primary search space for NAS techniques includes
various architectures that are different in design, the secondary search space includes subset of calculation methods for each possible architecture
inside the primary search space.
\par
Our focus in this work is on showing the effectiveness of applying Neural Twins Talk model over two different architectures for image captioning.
This is done in order to show that not only the Neural Twins Talk model performs better than similar models with one channel of attention, but also
to show that employing Neural Twins Talk over various models is achievable via utilizing alternative calculation methods.
\par
  \section{Related Work}\label{sec2}
\par
The closest related work to our work is NBT \cite{Lu2018NBT}.
NBT benefits from the improvements brought about by Bottom-up and Top-down Attention model \cite{Anderson2017up-down}
and provides us with information regarding the visual grounding of the generated words in the caption for
the input image. In NBT and Bottom-up and Top-down Attention \cite{Anderson2017up-down},
the authors pre-trained Faster-RCNN \cite{FasterRCNN} on COCO \cite{MS_COCO}
with a CNN backbone (Resnet101 \cite{Resnet}) trained on COCO and ImageNet \cite{imagenet_cvpr09} respectively,
creating the region proposal network \cite{FasterRCNN}. This mechanism served as the bottom-up attention.
On top of this object detector, there was a two layer top-down attention model that included
two Long Short Term Memory (LSTM) \cite{LSTM} units. The first LSTM acted as an attention LSTM while the second LSTM
unit acted as a language modeling unit to generate the captions for the input image.
This integrated mechanism produced an attentive language model that could be trained on embeddings
of the input caption and normalized coordinates from input image to generate the
captions for unseen images at test time \cite{Lu2018NBT,Anderson2017up-down}.
\par
Template generation, with slot filling \cite{Farhadi_2010,Li_2011,Kulkarni} was
among the first techniques used to solve the problem of automatic image caption generation.
Retrieval methods \cite{Hodosh_2015,Ordonez_NIPS2011,Sun_2015_ICCV} have also been used for generating image captions, these methods
retrieve a caption from caption data-base that best describes an image \cite{Hossain:2019:CSD:3303862.3295748}.
However, the most successful techniques have used deep learning models end-to-end for image caption generation.
Neural Twins Talk that we present in this paper uses a
twin cascaded attention model and is an instance of
a deep learning model end-to-end used for automatic image caption generation.
\par
Originally introduced by Sutskever et al. \cite{Sutskever_2014}, the encoder-decoder architecture divides the translation task into two parts.
The first portion performs the encoding process; in the context of image captioning, we could call it the feature extraction phase.
The second portion of this process is to pass the encoded features into another embedded space that acts as a decoder for generating the output sequence.
\par
Inspired by the encoder-decoder architecture \cite{Sutskever_2014},
an early work on image captioning using deep learning models by Kiors et al. \cite{kiros2014_1},
employed convolutional layers alongside a multi-modal
log bilinear LSTM to map 
visual and textual features in a shared multi-modal space.
\par
Karpathy et al. \cite{Karpathy_2014} used a method similar to
Kiros et al. \cite{kiros2014_1}, but improved the model
by modifying the way it generated embeddings for sentences and visual
features in multi-modal space by the LSTM unit.
``Show and tell'', introduced by Vinyals et al. \cite{Vinyals_2015_CVPR}, illustrated the fact that deep learning models could
handle the task of image captioning. They used simple CNN models such as AlexNet \cite{AlexNet} and VGG \cite{VGG} for feature extraction
and they used an LSTM as a language modeling unit. 
``Show Attend and Tell'' \cite{Vinyals_2015_CVPR} was one of the most interesting deep image captioning
models that demonstrated the usefulness of attention mechanisms in the context of image captioning.
Irrespective of the sub-architecture used for the encoder and decoder parts of the model,
the general idea is that an encoder extracts features and passes them to a
decoder for further analysis. Convolutional architectures \cite{AlexNet,VGG,Resnet,Resnext} and attention
mechanisms \cite{Famous_Cho_etal,Xu2,Jia_2015_ICCV}
have almost equally contributed to the success of image and video captioning and visual question answering tasks.
\par
Early deep learning methods for image captioning used straightforward convolutional architectures
that operate upon the entire image to extract the visual
features that is the encoder part of the deep image captioning model \cite{Karpathy_2014,Donahue_2015_CVPR,Xu,Xu2}.
Anderson et al. \cite{Anderson2017up-down} introduced the Bottom-Up and Top-Down Attention via using object detectors rather than straightforward convolutional networks in deep image captioning models.
Their idea was to make the model attend to different regions of the image to find
salient objects among proposed regions.
They used Faster R-CNN \cite{FasterRCNN}, which is a faster implementation of previous work called Fast R-CNN \cite{FastRCNN}. 
\par
Faster R-CNN \cite{FasterRCNN} uses a region proposal network in between each convolutional layer.
A bounding box with its coordinates is
proposed by the region proposal network. At a convolutional layer, the model refines the coordinates of these
regions and keeps adding more bounding boxes as it discovers more objects.
An interesting characteristic of Faster R-CNN is that it can use various convolutional architectures
as backbone for extracting visual features of the input image.
Therefore it is able to use any common architecture such as Alex-net \cite{AlexNet}
or Res-Nets \cite{Resnet} or the VGG model \cite{VGG}. 
\par
There are two main paradigms for using deep learning for generating captions from image features.
The first paradigm uses
the input image as a whole scene for generating captions \cite{Karpathy_2014,Vinyals_2015_CVPR}.
The second paradigm is the method of using important regions in the image similar to dense captioning \cite{DenseCap_2016} 
that generates captions for different regions and creates a final caption by
combining the generated captions for those regions \cite{Hossain:2019:CSD:3303862.3295748}. 
Baby Talk \cite{Kulkarni} was among the most successful slot filling models for image captioning. This work did not benefit from deep learning.
Baby Talk \cite{Kulkarni} produces
sentence templates that contain slots that could be filled in with the names of
the objects found in the image.
The problem with this approach was that the sentences created by this model were not able to show that they were
created by a model that is fully aware of how we humans tend to speak most of the time.
For this reason, the authors of NBT \cite{Lu2018NBT} introduced a novel framework for image captioning that combined
the earlier slot filling strategies with deep learning models that were able to handle language modeling.
This effective combination served as a foundation for NBT and also for our proposed attention model in this work.
\par
Lu et al. \cite{Lu2017KnowingWT} introduced a new attentive language model using a visual sentinel that
could attend over different parts of the input image and sentence embeddings. In this work,
they proposed the idea of adaptive attention that learned which regions were more important over time by learning the
joint relationship between captions from the training set and visual features from region detections. This idea was later used in NBT
to create a distinction between visual words and textual words in the caption sentence.
Pointer networks \cite{Oriol_2015_Ptr} were used in NBT in order to let the model
adaptively select the important region from the ``RoI align layer'' \cite{FasterRCNN}.
\par
  
\section{Methodology}\label{sec3}
\par
Without modifying the encoder part of the model (bottom-up attention), we only modify the decoder part of NBT (top-down attention) in order to show that twin cascaded
attention models are effective in making deep networks deploying LSTMs and attention mechanisms perform better.
In order to perform fair comparison, we
use the same training details such as the number of epochs and the batch size used for training the models.
We use the same object detector results to preserve the network
configurations in our experiments. The object detector used in our work was trained on COCO with a
ResNet-101 \cite{Resnet} as CNN backbone that was pre-trained on ImageNet \cite{jjNBT}.
\par
Similar to NBT \cite{Lu2018NBT}, given the input image, we find the parameters for the network to maximize the likelihood of correct caption for the given image.
Given an image sentence pair, while training the model, the goal of the model is to learn which words
in the ground truth caption can successfully be grounded into some region in the image. 
We maximize the $log$ likelihood of the correct caption using the summation of the joint probability of the given image and sentence pair. 
\par
By applying the chain rule, the joint probability distribution is obtained in terms of a sequence of
tokens in the caption generated by the model as the product of the probability of current token $y_t$ and all previously generated
tokens in the caption.
\par
A ``visual sentinel'' is used in NTT, similar to NBT \cite{Lu2018NBT}, to indicate if the current generated token should be a word describing
a region in the image or a word that creates the template sentence.
Given this new variable that acts as a default region sentinel, the
probability of token $y_t$ given an image and all previously generated tokens is computed.
This includes the computation of the joint probability
of the visual sentinel and the probability of current token $y_t$, multiplied by the probability of visual sentinel
based on the probability of input image and all previously generated tokens.

\par
\begin{figure}[t!]
    \centering
    \includegraphics[width=\linewidth, height=10cm, frame={\fboxrule} {-\fboxrule}]{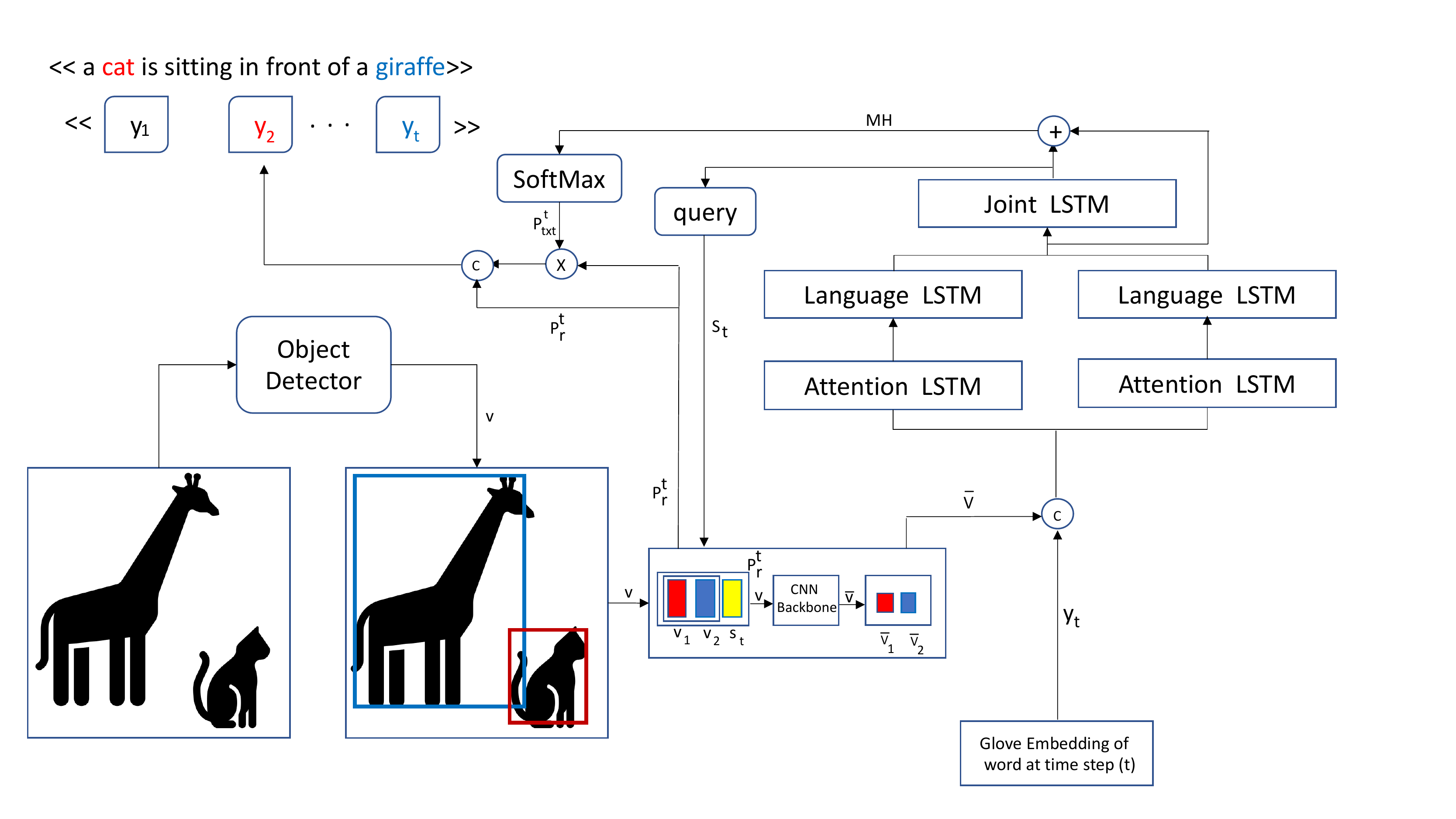}
    \caption{NTT framework from a general point of view. The main difference between NTT and
    NBT is that the language LSTMs in NTT receive their hypothesis and context vectors from their lower level attention LSTMs, rather than having their own vectors.
    Similarly the joint LSTM in NTT receives the hypothesis and context vectors from the lower level language LSTMs.
    This is explained in Section \ref{3.3}}
    \label{fig3}
\end{figure}
\subsection{Template Generation \& Refinement}\label{3.1}
\par
In the NTT model, we employ two channels of attention.
This causes the joint LSTM and language LSTMs to become capable of
refining the slots in an ensemble manner. We explain this general framework in Fig. \ref{fig3}.
At each time step using the default region sentinel, we determine if we need to use the
meta hypothesis vector for the textual word that creates the template sentence
or if we need to use the pointers for sub-categories and
plurality of the word to be placed inside a slot in the template sentence.
A Pointer Network \cite{Oriol_2015_Ptr} is used to create the slots in the template sentence \cite{Lu2018NBT}.
We modify the way the pointing vector is computed in NBT. We want to make sure the pointing vector is constructed
based on the hypothesis vector of the joint LSTM in our model instead of the language LSTM in NBT. At each
iteration, we calculate the current hypothesis of an RNN as
$h_t = RNN (x_t,h_{t-1})$. At each time step, $x_t$ is the ground truth token while testing, and
is the sampled token $y_{t-1}$ while training.
Consider $v_t\in \mathbb{R}^{d\times1}$ as the region feature vector for a region of
interest. Thus the pointing vector is calculated as the following.
\par
\begin{equation}\label{eq4}
    u_i^t=w_h^T \tanh(\textbf{W}_v\textbf{v}_t+\textbf{W}_z\textbf{h}_t^5)
\end{equation}
\begin{equation}\label{eq5}
    P_{ri}^t = softmax(u^t)
\end{equation}
\par
In Eq. \ref{eq4}, $\textbf{W}_v$ and $\textbf{W}_z$ are parameters to be learned by the model
and $\textbf{h}_t^5$ denotes the hypothesis vector of the joint language LSTM that connects the two channels of attention.
The visual sentinel in NTT is achieved by applying a gate when the RNN
is LSTM \cite{LSTM}, as explained in Eq. \ref{eq6} and Eq. \ref{eq7}.
\par
\begin{equation}\label{eq6}
    g_t = \sigma(\textbf{W}_x\textbf{x}_t+\textbf{W}_h\textbf{h}_{t-1}^5)
\end{equation}
\begin{equation}\label{eq7}
    s_t = g_t \odot tanh (c_t^5)
\end{equation}
\par
In Eq. \ref{eq7}, $c_t^5$ is the context of the joint LSTM at each time step $t$. In Eq. \ref{eq6}, $\textbf{W}_x$ and $\textbf{W}_h$ are weight parameters
to be learned and $\sigma$ denotes the sigmoid logistic function and
$\odot$ is the element-wise product and $x_t$ is the shared LSTM input at time step $t$. By modifying Eq. \ref{eq5} and
infusing the visual sentinel into that equation, we get the probability
distribution over the regions in the image as the following.
\par
\begin{eqnarray}\label{eq8}
    P_r^t= softmax([u^t;w_h^T tanh(\textbf{W}_s\textbf{s}_t+\textbf{W}_z\textbf{h}_t^5)]) 
\end{eqnarray}
\par
In Eq. \ref{eq8}, $\textbf{W}_s$ and $\textbf{W}_z$ are parameters and $P_r^t$ is the probability
distribution over the visual sentinel and grounding regions.
Next, we feed the meta hypothesis vector into a softmax layer. This is done in order to obtain
the probability of textual words regarding the visual features in the image,
and all previously generated words, and the visual sentinel as the following.
\par
\begin{equation}\label{eq9}
    P_{txt}^t = softmax(\textbf{W}_q\textbf{MH}_t)
\end{equation}
\par
In Eq. \ref{eq9}, $W_q\in \mathbb{R}^{S\times d}$ and $d$ is the hidden state size of
RNN and $S$ is the size of the textual vocabulary. In Section \ref{3.3}, we explain how $MH$ is calculated.
Infusing Eq. \ref{eq9} and the probability of default region sentinel based on the previous tokens into the probability of textual word in the template sentence based
on the default region sentinel
generates the probability of generating a word in the template sentence.
\par
Slot filling is performed on the region of interests. The convolutional feature maps
pooled from the selected RoI are sent to the attention LSTM; these feature maps are then processed by the
attention network before being passed to the language LSTM for template generation and refinement.
Using two single layer feed-forward networks with Relu
activation function denoted as $R(.)$,
we calculate the probability for plurality and fine grained sub-category class as the following.
\par
\begin{equation}\label{eq10}
    P_p^t = softmax(\textbf{W}_p R_b([v_t;h_t^5]))
\end{equation}
\begin{equation}\label{eq11}
    P_{sc}^t = softmax(\textbf{U}^T\textbf{W}_{sc} R_g([v_t;h_t^5]))
\end{equation}
\par
In Eq. \ref{eq10} and Eq. \ref{eq11}, $\textbf{U}$ is the vector of embedding of the word for the sub-category that fills the slots,
and $W_p$ and $W_{sc}$ are weight parameters to be learned.
The last phase of caption template refinement is 
to consider $P_p$ that is the probability for plurality and $P_{sc}$ that
is the probability for the sub-category for the words that are going to fill the slots in the template sentence.
Eq. \ref{eq4} - Eq. \ref{eq11} are similar to how they are presented in NBT, except that
instead of using the hypothesis and context vectors coming from the single language LSTM in NBT,
we use the meta hypothesis vector and hypothesis and context vectors coming from the joint LSTM in our model.
\par
\subsection{Loss Function \& Training}\label{3.2}
\par
We minimize Cross-Entropy loss function, also used in NBT.
Regarding visual word extraction, detection model, region feature extraction, previously proposed attentive language model
and other implementation details,
we encourage the readers to refer to Neural Baby Talk \cite{Lu2018NBT}.
\par
We train both models on four Nvidia 1080ti GPU cards and we use batch size of 100. 
We retrain the NBT model and our proposed model (NTT) both from scratch on COCO
using the split for this dataset provided by Karpathy \cite{Karpathy_2017} and the novel and robust image captioning splits
provided by Hendricks et al. \cite{Hendricks_2016_CVPR} and Lu. et al. \cite{Lu2018NBT,jjNBT}. In our experiments, we use a consistent beam size of 3.
\par
The difference between our method of training and the original
one in NBT is that instead
of using a ResNet101 \cite{Resnet} trained on COCO, we use another version of this CNN backbone trained
on ImageNet \cite{imagenet_cvpr09} for region feature extraction.
We only fine-tune the last layer of the CNN backbone for the region feature extraction phase while training both models.
We train both models for 50 epochs with Adam optimizer \cite{Adam} and we anneal
the learning rate every $3$ epochs by a factor of $0.8$. Similar to NBT,
we use Glove embeddings \cite{Glove}
for creating word embeddings from the words in the ground truth captions. 
\par

\subsection{Proposed Attention Model}\label{3.3}
\par
The general framework of NTT was explained in Fig. \ref{fig3}.
Inspired by residual learning \cite{Resnet} and multi-head attention \cite{NIPS2017_7181}, we create parallel attention channels and
introduce cascaded adaptive gates in our model to employ residual connections between parallel channels.
This is why we refer to our proposed decoder as twin cascaded attention decoder.
\par
\begin{figure}[t!]
    \centering
    \includegraphics[width=\linewidth, height=10cm,frame={\fboxrule} {-\fboxrule}]{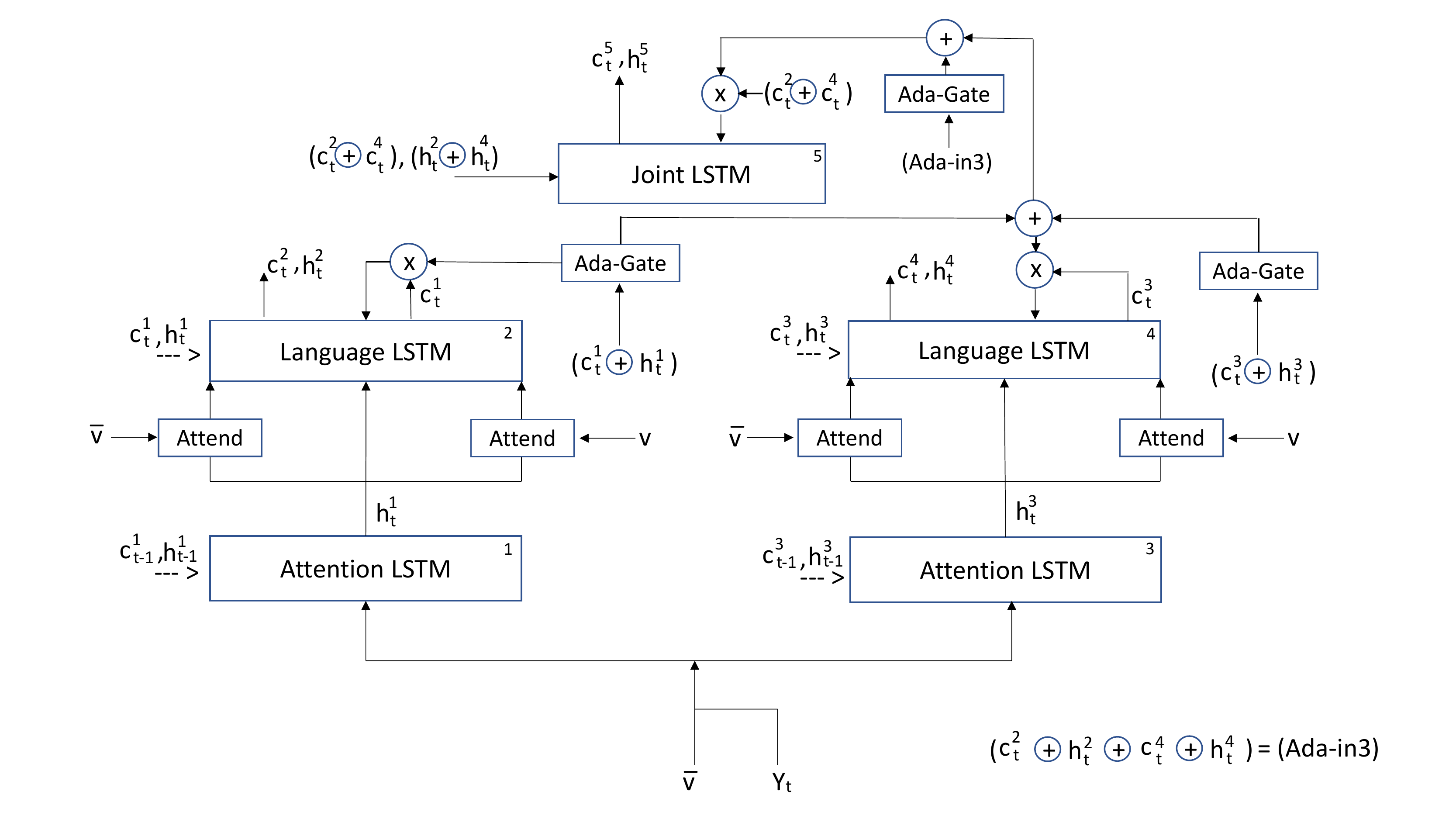}
    \caption{Twin cascaded attention model proposed in our work. The cascaded adaptive gates connect the parallel attention channels.
    At each time step, language LSTMs and the joint LSTM receive the necessary information from their lower level
    LSTMs.}
    \label{fig4}
\end{figure}

The shared input of both attention LSTMs in our proposed model is the concatenation of
an embedding of token $y_t$ and the set of convolutional features from region proposals $\bar{V}$. At each time step $t$, the context and hypothesis vectors
of the attention LSTM in the first channel of the twin cascaded decoder on the left side are
passed to the language LSTM in that channel. Similarly, the context and hypothesis vectors
of the attention LSTM in the second channel of the twin cascaded decoder on the right are
passed to the language LSTM in that channel. Then, the context vector coming from the language LSTM in the first channel is
added to the context vector coming from the language LSTM in the second channel
to form the context vector for the joint LSTM. We follow a similar strategy
to construct the hypothesis vector for the joint LSTM. This is shown in
Fig. \ref{fig4}. We believe this causes the joint LSTM to perform
the attention once again, this time, given the context and hypothesis vectors of the first language LSTM
on the left side of decoder and the second language LSTM on the right side to learn to perform the attention better.
\par
\par
After the hypothesis vectors of the attention LSTMs are computed, they are passed to an attention network.
A replica of the attention network is present in each channel. This implies that
the first attention network in the top-down channel calculates the attention distribution
over $V$ set of region features (proposals), and then the second attention network in the same
channel calculates the attention distribution over $\bar{V}$, the set of convolutional features from region proposals.
\par
By adding adaptive gates, and performing the residual
learning upon these gates, the attention is successfully passed to lower levels of the decoder.
This causes the language LSTMs to work in an ensemble manner.
The language LSTMs keep helping one another in generating and refining the template sentences.
This happens in a cascaded manner as shown in Fig. \ref{fig4}.
At each time step, the language LSTMs in the parallel attention channels receive the hypothesis and context
vectors from their lower level attention LSTM in the attention channel they reside in.
Similarly, the joint LSTM receives the necessary information from the language LSTMs in
each attention channel.
This eliminates the need for gathering the information at each time
step from the language LSTMs and the joint LSTM, which reduces memory usage. Therefore,
the extra memory required by NTT in comparison with NBT, is only around $15\%$, and most of this increase in memory usage
is contributed to feed-forward networks employed in cascaded adaptive gates.
We add the context and hypothesis vectors of each language LSTM in the twin attention channels with
each other, and we pass it to the joint language LSTM in our model.
In a sense, it is a summation of the vectors for all the previous language LSTMs.
The attention distribution over $V$ set of region features is calculated as explained in Eq. \ref{eq12}.
\par
\begin{equation}\label{eq12}
    \begin{split}
        \beta_t^1 &= \textbf{W}_\beta^T (\textbf{W}_v\textbf{V}+(\textbf{W}_h\textbf{h}_t^1) \mathbf{1}^T) \\
        \alpha_t^1 &= softmax(\beta_t^1) \\
        \beta_t^2 &= \textbf{W}_\beta^T (\textbf{W}_v\textbf{V}+(\textbf{W}_h\textbf{h}_t^3) \mathbf{1}^T) \\
        \alpha_t^2 &= softmax(\beta_t^2)
    \end{split}
\end{equation}
\par
In Eq. \ref{eq12}, ${W}_\beta^T$, $W_v$ and $W_h$ are weight parameters to be learned
by the model. These parameters are shared between the language LSTMs in the decoder
channels. The set of attention values $\alpha_t^1$ is received by
the language LSTM in the attention channel on the left side of the twin cascaded
decoder. Similarly, $\alpha_t^2$ is the set of attention values received by the language
LSTM in the attention channel on the right side of the twin cascaded decoder.
In total, this attention network is used four times in our model; once for $V$ set of region features, and
once for $\bar{V}$, the set of convolutional features from region proposals for the language LSTM in each attention channel.
\par
Next, we show how adaptive gates are calculated in our decoder. 
Given the input of a language LSTM in an attention channel of our decoder and the hypothesis vector coming form attention LSTM in the same attention channel,
we calculate the values for the adaptive gate in that channel. These gates are added to each other in a cascaded manner as shown in Fig. \ref{fig4}.
The values for the adaptive gate applied to the joint LSTM are obtained by including the previously cascaded adaptive gates in the attention channels and a mixture of
the input of language LSTMs attention channels.
This is shown in Eq. \ref{eq13}.
\par
\begin{equation}\label{eq13}
    \begin{split}
        AdaGate_1 &= \sigma(W_A^1(h_t^1 \oplus c_t^1)) \\
        AdaGate_2 &= \sigma(W_A^2(h_t^3 \oplus c_t^3)) \\
        AdaGate_2 &= AdaGate_2 \oplus AdaGate_1 \\
        AdaGate_3 &= \sigma(W_A^3(h_t^2 \oplus c_t^2 \oplus h_t^4 \oplus c_t^4)) \\
        AdaGate_3 &= AdaGate_3 \oplus AdaGate_2 \\
    \end{split}
\end{equation}
\par
In Eq. \ref{eq13}. $W_A^1$, $W_A^2$ and $W_A^3$ are the weight parameters for adaptive gates to be found by the model.
Each of these adaptive gates are applied to the context vector of their respective
language LSTM unit in the decoder. This is shown in Eq. \ref{eq14}.
We find that by adding the adaptive gates on each other, the model learns to
attend better on different parts of the input for each language LSTM.
\par
\begin{equation}\label{eq14}
    \begin{split}
        c_t^2 &= AdaGate_1 \odot c_t^2 \\
        c_t^4 &= AdaGate_2 \odot c_t^4 \\
        c_t^5 &= AdaGate_3 \odot c_t^5 \\
    \end{split}
\end{equation}
\par
The inputs of the final language LSTM in our decoder, which we refer to as the joint
language LSTM, are perhaps the most important parts of our decoder.
Considering that we have two top-down attention channels, we want another language LSTM that receives the output
of both of these channels ($h_t^2,h_t^4$) with their context vectors ($c_t^2,c_t^4$) jointly to
refine the generated caption one more time. Eq. \ref{eq15} and Eq. \ref{eq16} show how the inputs of joint language LSTM is created at each time step.
\par
\begin{equation}\label{eq15}
    \begin{split}
        h_t^5 &= h_t^2 \oplus h_t^4 \\
        c_t^5 &= c_t^2 \oplus c_t^4 \\
    \end{split}
\end{equation} 
\begin{equation}\label{eq16}
    \begin{split}
        LSTM_{in}^1 = [ \alpha_t^1 ; h_t^1] \\
        LSTM_{in}^2 = [ \alpha_t^2 ; h_t^3] \\
        LSTM_{in}^3 = LSTM_{in}^2 \oplus LSTM_{in}^1
    \end{split}
\end{equation}
\par
The final output is calculated based on $h_t^5$, $h_t^2$ and $h_t^4$, the hypothesis
vectors coming from the last language LSTM in the decoder module as well as
the other two language LSTMs in the attention channels.
Note that we refer to the concatenation of the hypothesis vector coming from the attention LSTM and
the output of the attention networks in each channel as language LSTM input and
we show it as $LSTM_{in}$. In Eq. \ref{eq16}, $LSTM_{in}^1$ denotes the input of language LSTM in the left side channel and $LSTM_{in}^2$
denotes the input of language LSTM in the right side channel of the decoder, similarly $LSTM_{in}^3$ denotes the input of the joint LSTM that is the
result of the element-wise addition between $LSTM_{in}^1$ and $LSTM_{in}^2$.
\par
The final output of our model comes from the language LSTMs in attention channels and the joint LSTM in our model.
The meta hypothesis is a summation of the output of
dropout layers applied to the hypothesis vectors coming form the language LSTMs
and joint LSTM. This is done by applying a dropout \cite{Hinton2014_dropout}
rate of 0.3 on the output of the language LSTM in the left-side attention channel.
Next, we apply a dropout rate of 0.7 on the language LSTM in the second
attention channel. Lastly, we apply a dropout rate of 0.8 on
the hypothesis vector coming from the joint LSTM that connects the two attention
channels with each other. We calculate the summation of the outputs of these dropout layers to create the final
hypothesis vector to form the meta hypothesis vector.
We show how the meta hypothesis vector is constructed in Eq. \ref{eq17}. 
\par
\begin{equation}\label{eq17}
    \begin{split}
        h_t^2 &= Dropout(h_t^2) : Rate=0.3\\
        h_t^4 &= Dropout(h_t^4) : Rate=0.7 \\
        h_t^5 &= Dropout(h_t^5) : Rate=0.8 \\
        MH_t &= h_t^2 \oplus h_t^4 \oplus h_t^5 \\
        MH_t &= Dropout(MH_t) : Rate=0.5
    \end{split}
\end{equation}
\par
In Eq. \ref{eq17}, $MH_t$ denotes the final output of our model, which we refer to as the meta hypothesis vector at time step $t$.
We use this vector to generate each word in the caption sentence at time step $t$. 
\par

\subsection{Alternative Calculations}\label{3.4}
\par
Bottom-up and Top-down attention (Up-Down Attention) model \cite{Anderson2017up-down}, which was published prior to Neural Baby Talk (NBT) \cite{Lu2018NBT},
was trained on image captioning task without visual grounding.
In NBT, for visual grounding we have to deal with template generation and refinement using slot filling.
This is done using two attention networks for a language LSTM. One of the attention networks is used for bounding box coordinates and the other one is used
for visual features extracted from the bounding boxes. In Up-Down Attention, the model only employs one attention network for the language LSTM
in order to calculate the attention values based on visual features and the hypothesis vector coming from the attention LSTM.
\par
We use Eq. \ref{eq12} in the same way it is defined in Neural Twins Talk utilized over Up-Down Attention.
Instead of using new sets of attention weights for bounding box coordinates and visual features as in NTT over NBt,
here we only use Eq. \ref{eq12} once in each attention channel that includes an attention LSTM, an attention network and a language LSTM, in order to perform attention over set of visual features.
\par
In order to show that Neural Twins Talk model can be applied on Up-Down Attention model \cite{Anderson2017up-down} we employ the same
expansion technique that was used for applying Neural Twins Talk on Neural Baby Talk model \cite{Lu2018NBT}. To be specific, the proposed attention model
in Neural Twins Talk used for enhancing Neural Baby Talk with more attention channels has an architecture identical to the one utilized over Up-Down Attention model.
\par
Instead of modifying the architecture of Neural Twins Talk we modify the way the
hypothesis and context vectors are gathered and distributed within LSTM units.
The original hypothesis and context gathering in Neural Twins Talk is novel in that instead of gathering the hypothesis and context vectors of LSTM units individually,
each language LSTM in each attention channel receives the hypothesis and context vectors from their corresponding lower level attention LSTMs.
This is also explained in Fig. \ref{fig4}.
\par
In order to achieve better performance when applying Neural Twins Talk on Up-Down Attention,
the hypothesis and context gathering and distribution method is modified.
Equations \ref{eq12} - \ref{eq17} are identically used in Neural Twins Talk when utilized over Up-Down Attention.
Instead of gathering the hypothesis and context vectors for attention LSTMs individually when applying Neural Twins Talk on Neural Baby Talk,
here we merge the hypothesis and context vectors coming from the attention LSTMs and the merged hypothesis and context vectors are passed to both
attention LSTMs for next iteration step. This is explained in the following equation, note that the LSTMs are assigned with the same numbers as they were
in Neural Twins Talk employed for Neural Baby Talk.
\par
\begin{equation}\label{eq18}
    \begin{split}
        LSTM^1(h^1_n,c^1_n) &= LSTM^1([h^1_{n-1}\oplus h^3_{n-1}], [c^1_{n-1}\oplus c^3_{n-1}]) \\
        LSTM^3(h^3_n,c^3_n) &= LSTM^3([h^1_{n-1}\oplus h^3_{n-1}], [c^1_{n-1}\oplus c^3_{n-1}]) \\
    \end{split}
\end{equation}
\par
Also instead of gathering the hypothesis and context vectors for language LSTMs in each attention channel from their lower level attention LSTMs, we
gather the information from the joint LSTM ($LSTM^5$), which conveys
the output hypothesis and context vectors coming from the joint LSTM to the lower level language LSTMs ($LSTM^2, LSTM^4$) for next iteration.
This is shown in the following equations.
\par
\begin{equation}\label{eq19}
    \begin{split}
        LSTM^2(h^2_n,c^2_n) &= LSTM^2(h^5_{n-1}, c^5_{n-1}) \\
        LSTM^4(h^4_n,c^4_n) &= LSTM^4(h^5_{n-1}, c^5_{n-1}) \\
    \end{split}
\end{equation}
\par
\begin{equation}\label{eq20}
    \begin{split}
        LSTM^5(h^5_n,c^5_n) &= LSTM^5([h^2_{n}\oplus h^4_{n}], [c^2_{n}\oplus c^4_{n}]) \\
    \end{split}
\end{equation}
\par
The joint LSTM ($LSTM^5$) receives the merged hypothesis and context vectors coming from language LSTMs ($LSTM^2, LSTM^4$) at each time step.
This is identical to how the hypothesis and context vectors for the joint LSTM were calculated when Neural Twins Talk was utilized over Neural Baby Talk.
This is explained in Eq. \ref{eq20}.
\par
Instead of modifying the dropout rates and how the adaptive gates are calculated,
we only modify the way the hypothesis and context vectors are gathered and distributed
in a joint manner from the attention LSTMs in the first level of attention channels, and from the joint LSTM and language LSTMs in a circular manner.
The inputs of language LSTMs and the joint LSTMs are identical to how they were calculated when Neural Twins Talk was utilized over Neural Baby Talk.
This is shown in Eq. \ref{eq16}.
\par
In NTT on NBT model, the joint LSTM, which receives the merged hypothesis and context vectors from the lower level language LSTMs, plays a key role of
acting as a bridge between attention channels in this model.
\par
In order to reveal the importance of having a joint LSTM in NTT we follow the modified alternative
calculation method and at the same time we remove the joint LSTM and instead we add a third attention channel.
\par
In this modified version of NTT on Up-Down Attention model
we want to investigate the tradeoff between removing the joint LSTM and instead adding another parallel attention. The hypothesis and context vectors of
all attention LSTMs are merged and fed back to each attention LSTM individually. Similarly, The hypothesis and context vectors of
all language LSTMs are merged and fed back to each language LSTM individually without having the joint LSTM.
The following equations show how the hypothesis and context vectors are calculated in this version of NTT on Up-down.
Note that all attention LSTMs are denoted with odd numbers in Eq. \ref{eq21} and all
language LSTMs are denoted with even numbers in Eq. \ref{eq22}.
\par
\begin{equation}\label{eq21}
    \begin{split}
        LSTM^1(h^1_n,c^1_n) &= LSTM^1([h^1_{n-1}\oplus h^3_{n-1}\oplus h^5_{n-1}], [c^1_{n-1}\oplus c^3_{n-1}\oplus c^5_{n-1}]) \\
        LSTM^3(h^3_n,c^3_n) &= LSTM^3([h^1_{n-1}\oplus h^3_{n-1}\oplus h^5_{n-1}], [c^1_{n-1}\oplus c^3_{n-1}\oplus c^5_{n-1}]) \\
        LSTM^5(h^5_n,c^5_n) &= LSTM^5([h^1_{n-1}\oplus h^3_{n-1}\oplus h^5_{n-1}], [c^1_{n-1}\oplus c^3_{n-1}\oplus c^5_{n-1}]) \\
    \end{split}
\end{equation}
\par
\begin{equation}\label{eq22}
    \begin{split}
        LSTM^2(h^2_n,c^2_n) &= LSTM^2([h^2_{n-1}\oplus h^4_{n-1}\oplus h^6_{n-1}], [c^2_{n-1}\oplus c^4_{n-1}\oplus c^6_{n-1}]) \\
        LSTM^4(h^4_n,c^4_n) &= LSTM^4([h^2_{n-1}\oplus h^4_{n-1}\oplus h^6_{n-1}], [c^2_{n-1}\oplus c^4_{n-1}\oplus c^6_{n-1}]) \\
        LSTM^6(h^6_n,c^6_n) &= LSTM^6([h^2_{n-1}\oplus h^4_{n-1}\oplus h^6_{n-1}], [c^2_{n-1}\oplus c^4_{n-1}\oplus c^6_{n-1}]) \\
    \end{split}
\end{equation}
\par
In the modified version of NTT on Up-Down Attention model, which was explained above, each attention channel has an attention LSTM followed by a language LSTM.
This version of NTT on Up-Down Attention only benefits from a third parallel attention channel
in comparison with the first version explained in the beginning of section \ref{3.4}.
\par
The final meta hypothesis vector that is used as the output of the model is calculated similar to NTT on NBT. The difference here is that
the joint LSTM is removed and replaced with the third language LSTM in the third attention channel. For simplicity the third language LSTM was
denoted as $LSTM^6$. Considering this change, the meta hypothesis vector calculation method is identical to the one explained in Eq. \ref{eq17}.
\par
The inputs of language LSTMs for the first language LSTM and second language LSTM are identical to how they were explained in Eq. \ref{eq16}.
For the third language LSTM ($LSTM^6$), similar to the first and second language LSTMs,
the input consists of a concatenation of attention values calculated from visual
features and the hypothesis vector coming from the corresponding lower level attention LSTM ($LSTM^5$).
\par
Performing experiments with these two models that follow the alternative calculation method of
hypothesis and context gathering, should reveal whether having a joint LSTM with two parallel attentions
is more effective for performance gain or having a third parallel channel without a joint LSTM instead.
At the same time we could investigate if alternative calculations are effective in performance gain when employed on the same neural expansion algorithm (NTT) over
different models such as NBT or Up-Down Attention.
The results of experiments with both versions of NTT on Up-Down Attention using alternative calculation methods is discussed in Section \ref{sec4.2}.
\par
  
\section{Discussion \& Results}\label{sec4}
\par
In this section we discuss the results of experiments for Neural Twins Talk applied on Neural Baby Talk \cite{Lu2018NBT} in Section \ref{sec4.1} and then we
discuss the results of experiments for Neural Twins Talk applied on Bottom-up and Top-down Attention \cite{Anderson2017up-down} in Section \ref{sec4.2}.

\subsection{NTT over NBT}\label{sec4.1}
\par
We report the results of our experiments for three different splits
on COCO. The first split is provided to us by Karpathy et al. \cite{Karpathy_2015_CVPR}; this split is commonly used for image captioning using deep learning models.
The more challenging splits proposed by Hendricks et
al. \cite{Hendricks_2016_CVPR} and Lu et al. \cite{Lu2018NBT} are Novel and Robust splits.
\par
\begin{figure*}[!ht]\centering
    \includegraphics[width=8cm, height=5cm, frame={\fboxrule} {-\fboxrule}]{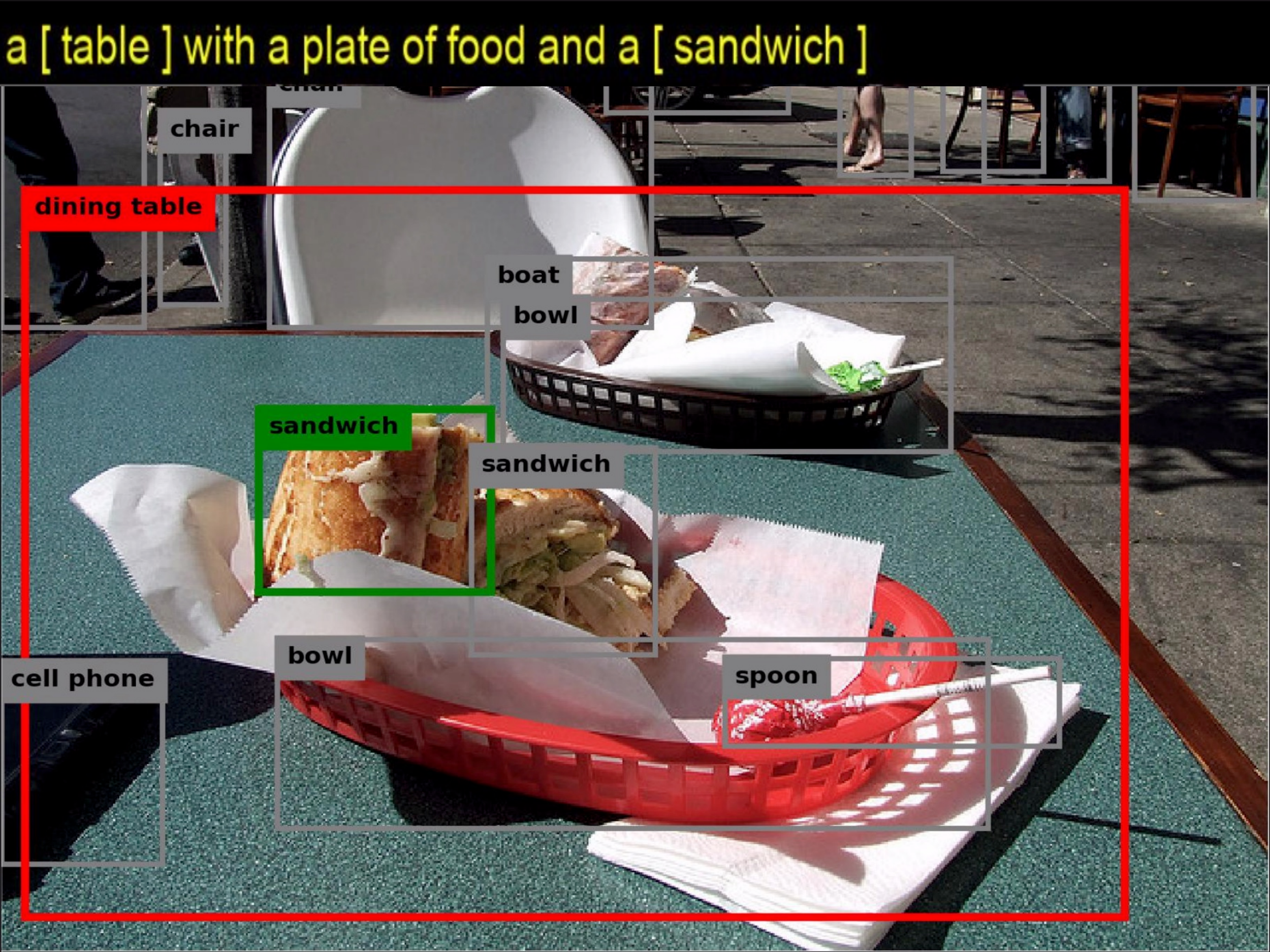}
    \includegraphics[width=8cm, height=5cm, frame={\fboxrule} {-\fboxrule}]{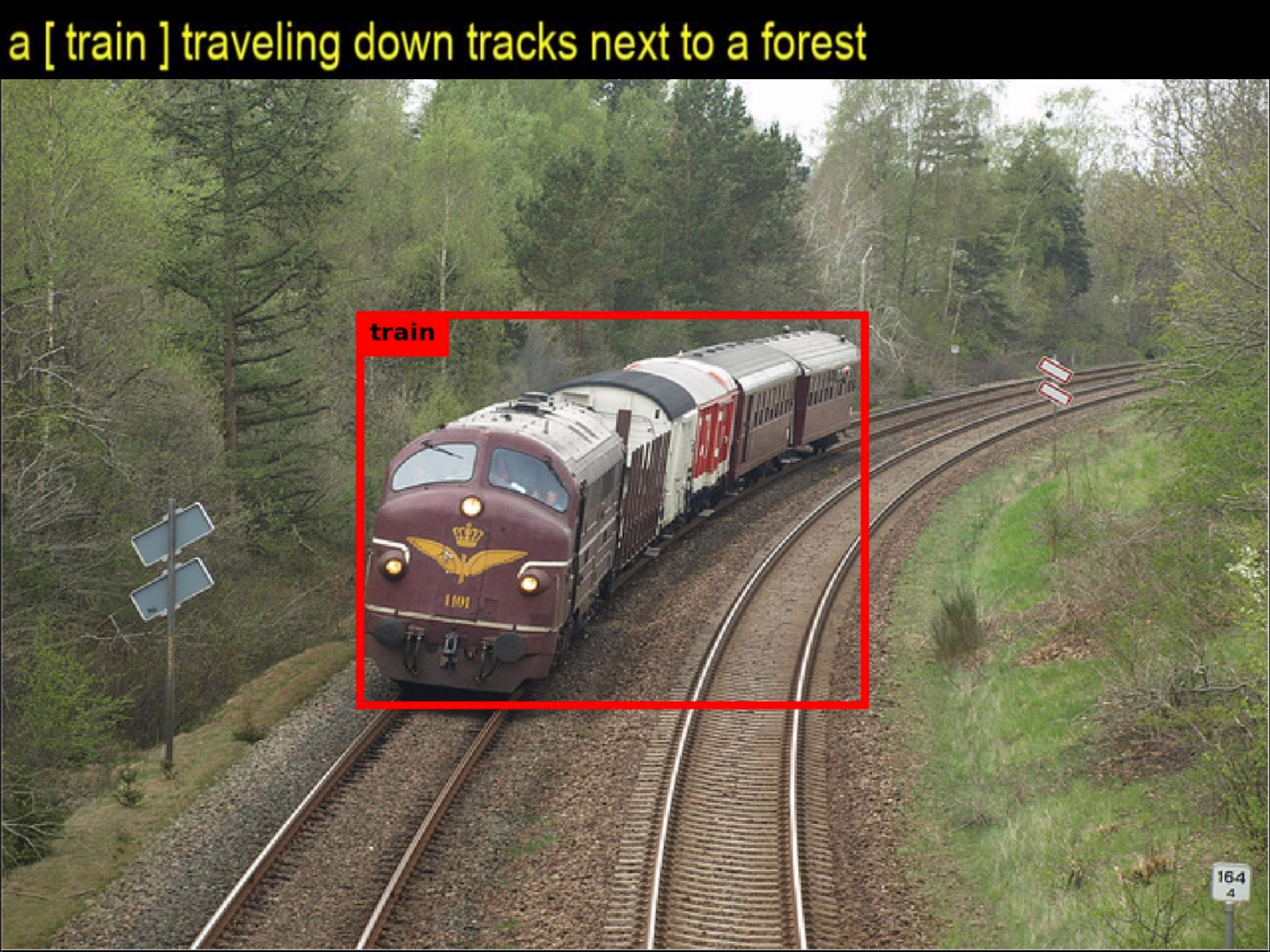}
    \includegraphics[width=16.1cm, height=8cm, frame={\fboxrule} {-\fboxrule}]{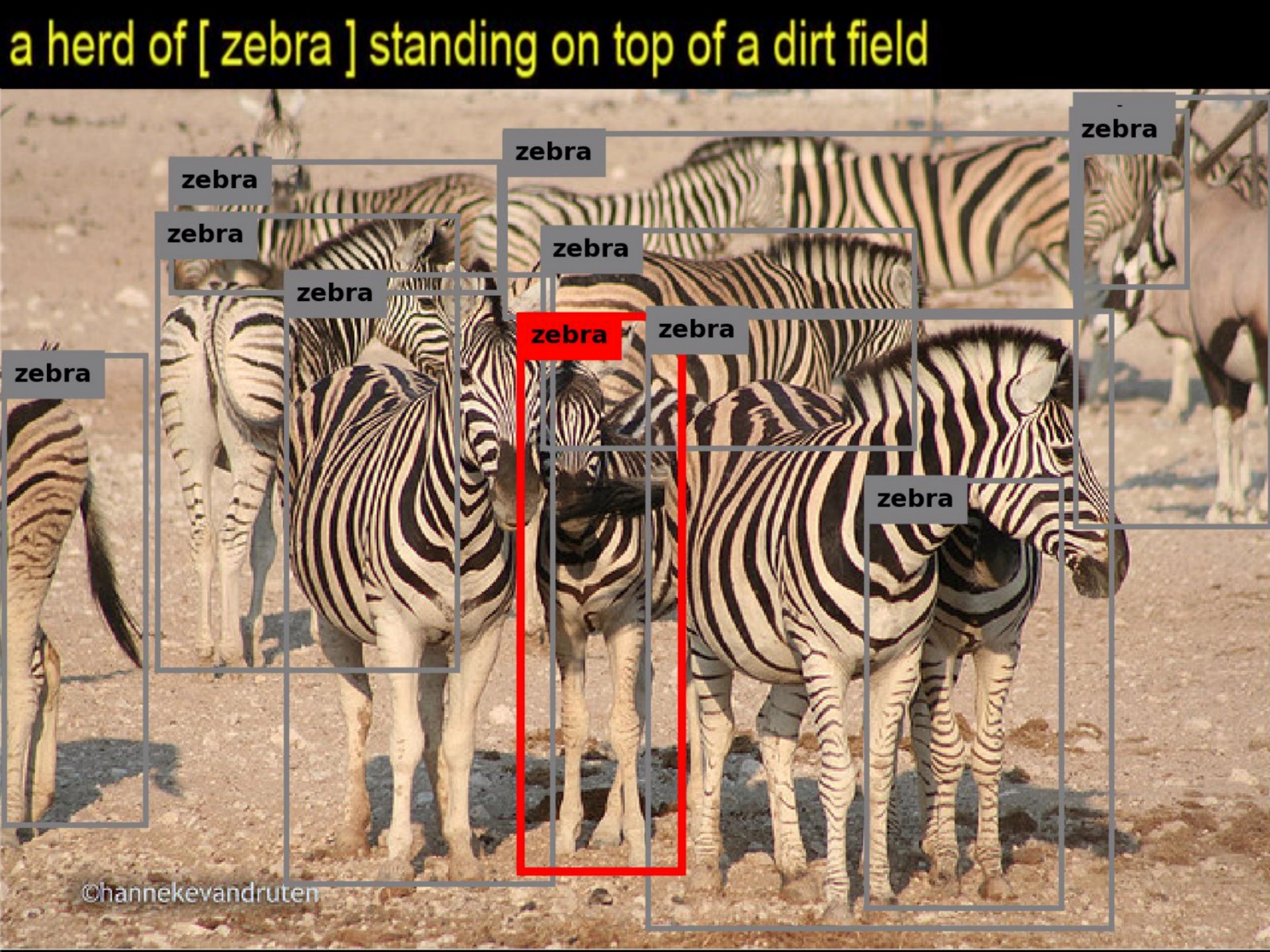}
    \caption{Examples of the results of our proposed attention model on the Karpathy's split \cite{Karpathy_2015_CVPR}.
    The results from this split show that the generated captions are relevant to the image and slots
    are filled successfully. The result is a rich caption that explains the scene successfully.}
    \label{fig5}
\end{figure*}
\begin{figure*}[!ht]\centering
    \includegraphics[width=8cm, height=5cm, frame={\fboxrule} {-\fboxrule}]{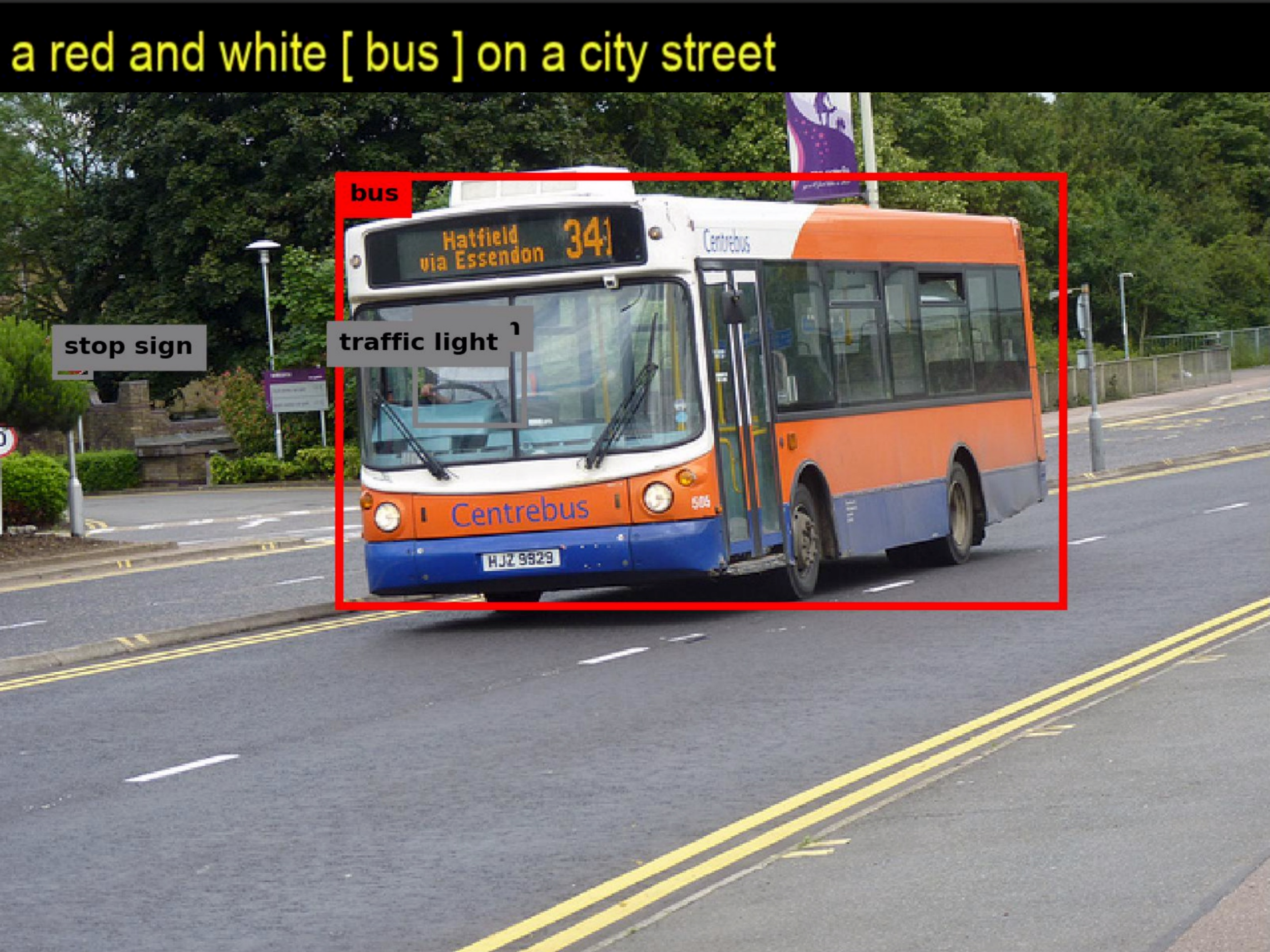}
    \includegraphics[width=8cm, height=5cm, frame={\fboxrule} {-\fboxrule}]{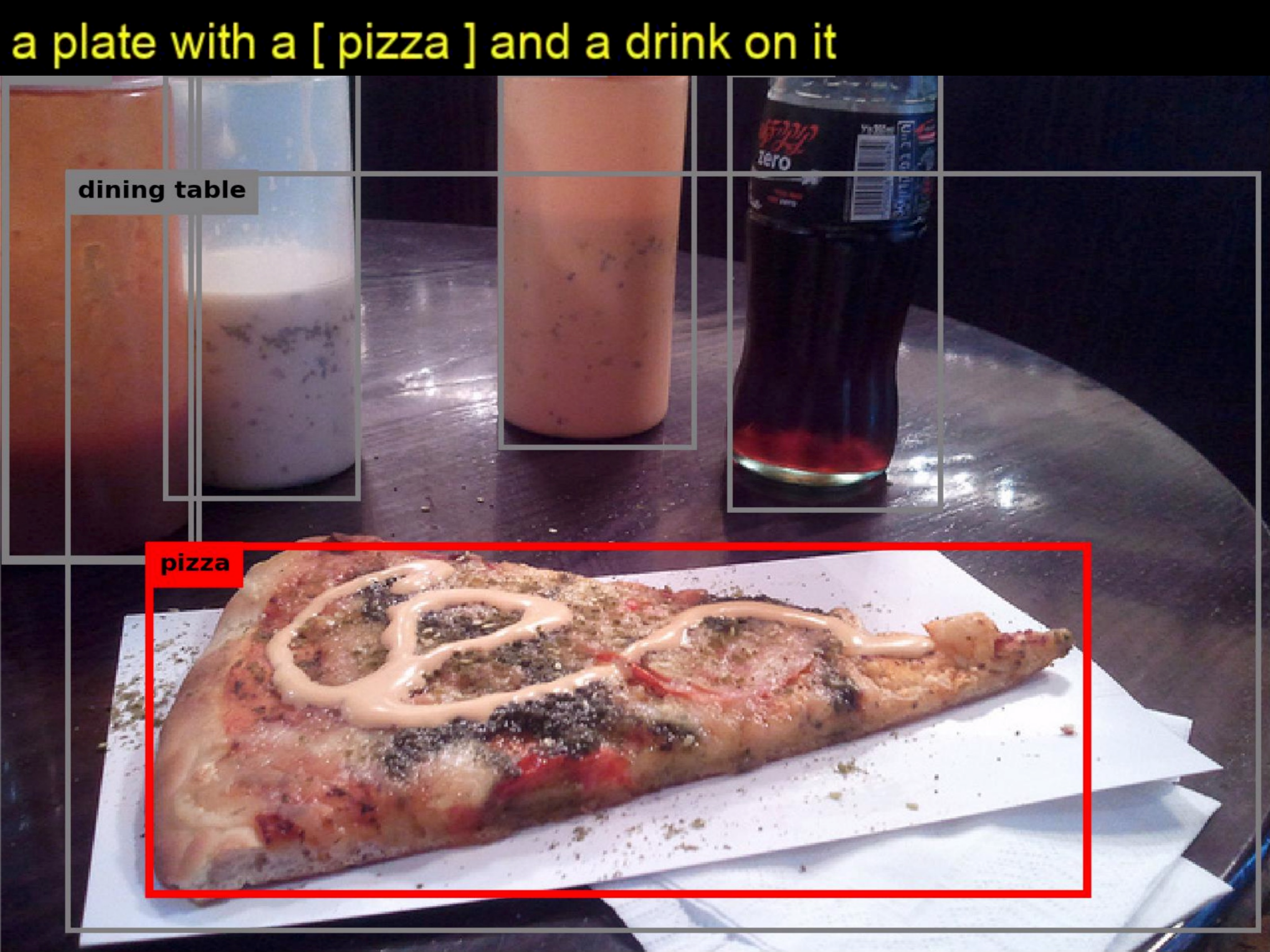}
    \includegraphics[width=16.1cm, height=8cm, frame={\fboxrule} {-\fboxrule}]{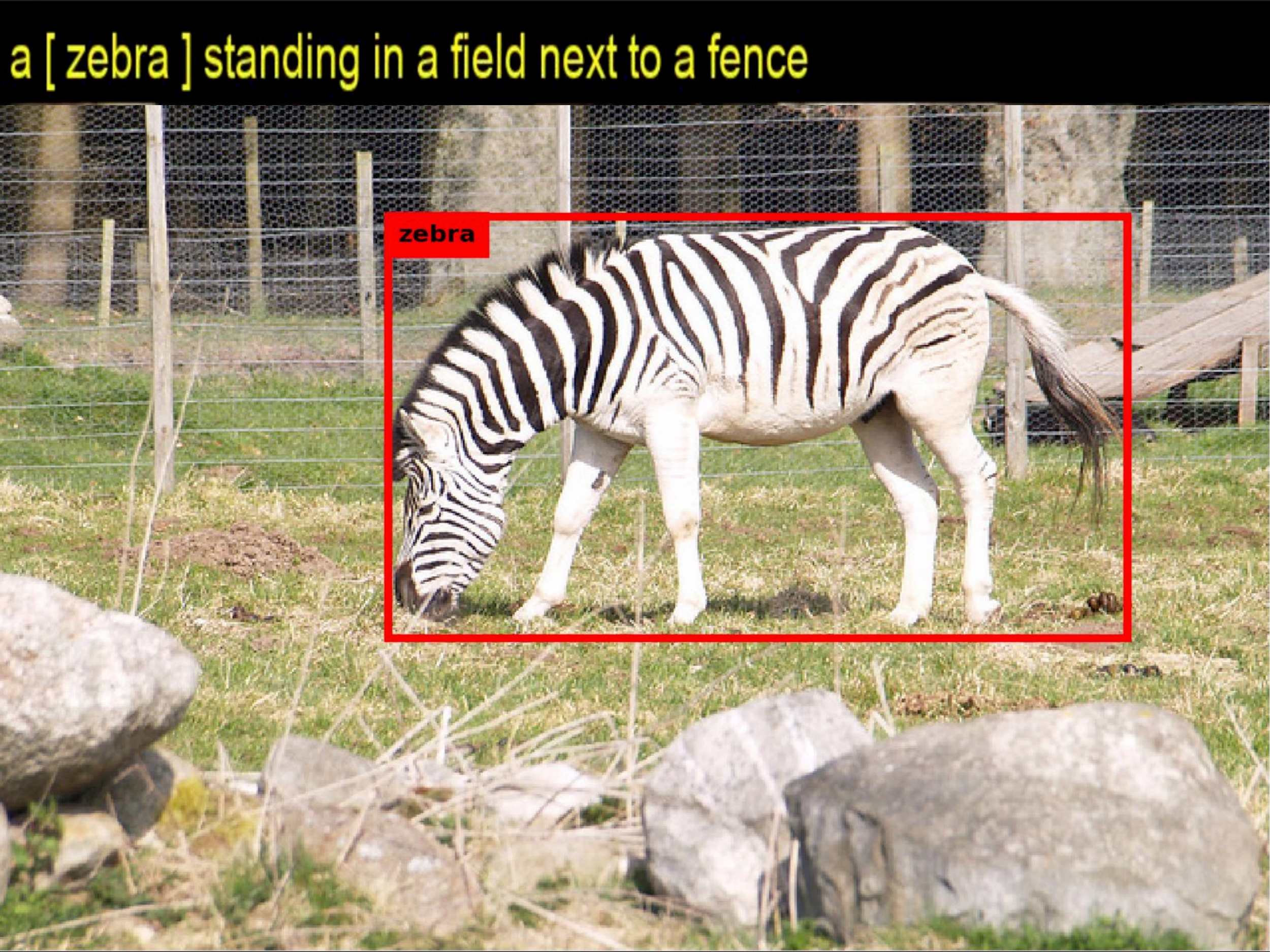}
    \caption{Examples of the results of our proposed attention model on the Novel
    split \cite{Hendricks_2016_CVPR}. These results show that the model is able to
    generate captions for in-domain images that include objects
    that were excluded from the train set. The objects are successfully detected and fill the slots
    in the caption.}
    \label{fig6}
\end{figure*}
\begin{figure*}[!ht]\centering
    \includegraphics[width=8cm, height=5cm, frame={\fboxrule} {-\fboxrule}]{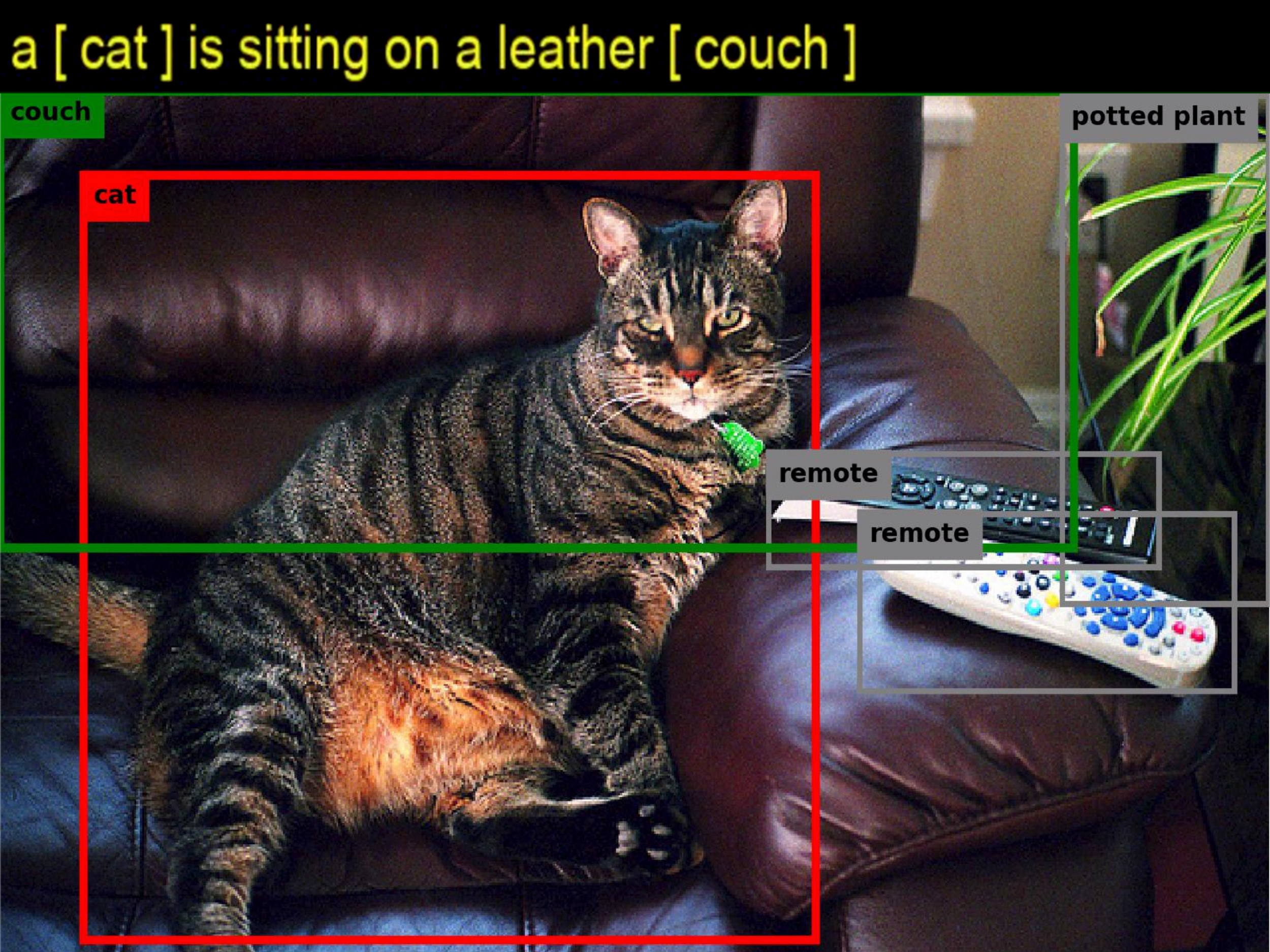}
    \includegraphics[width=8cm, height=5cm, frame={\fboxrule} {-\fboxrule}]{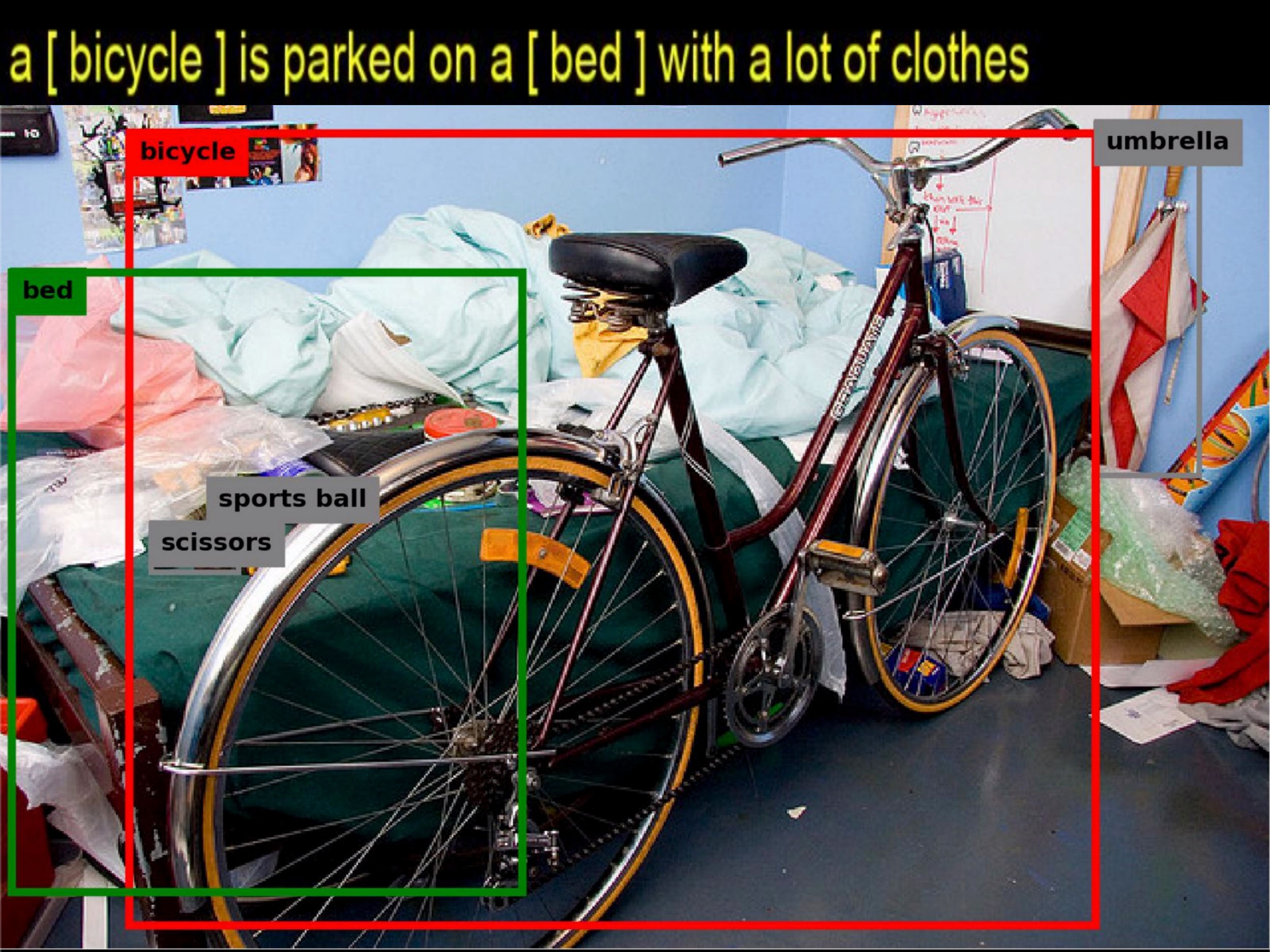}
    \includegraphics[width=16.1cm, height=8cm, frame={\fboxrule} {-\fboxrule}]{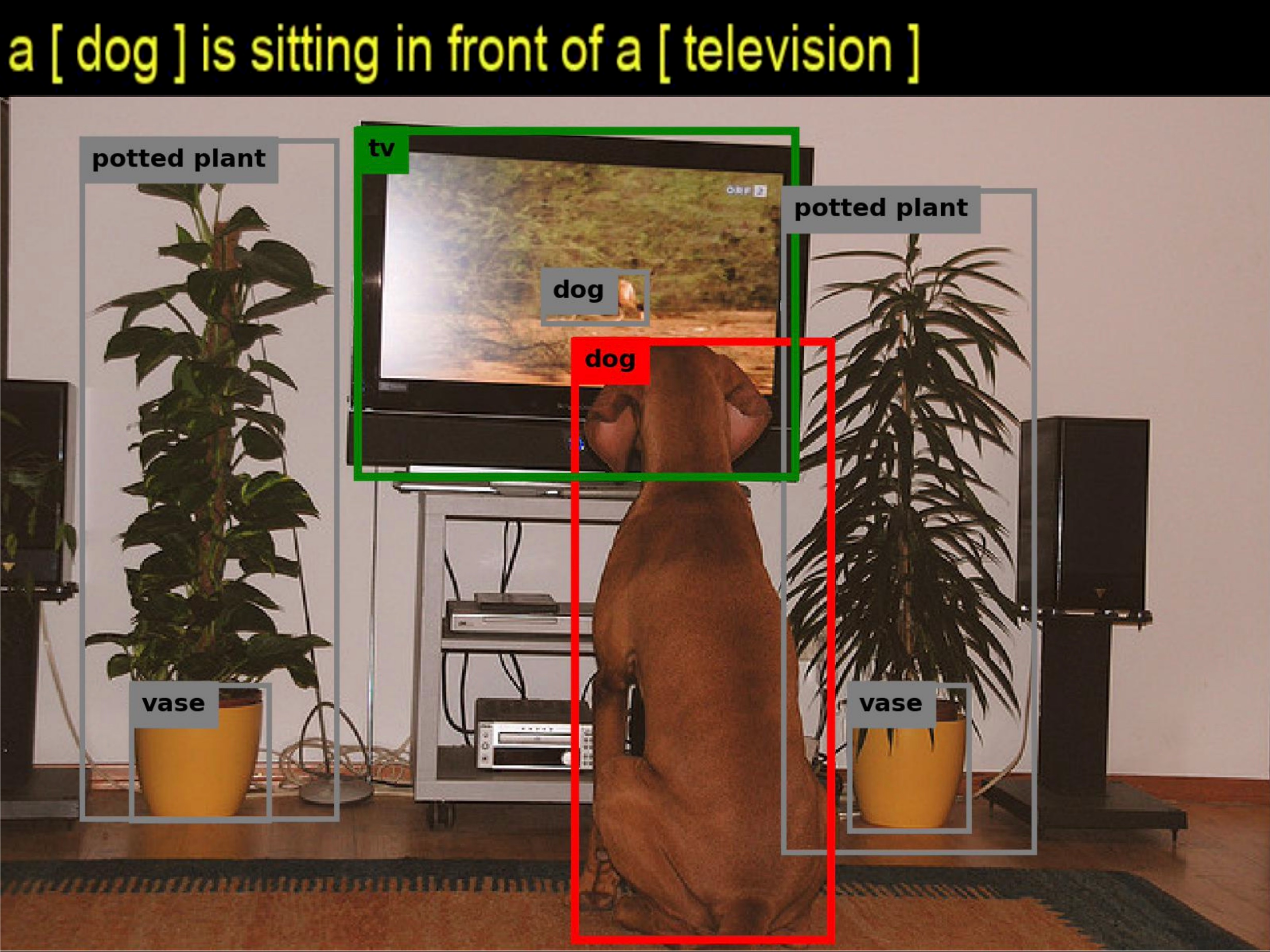}
    \caption{Examples of the results of our proposed attention model on the
    Robust split \cite{Lu2018NBT}. These results show that the model is able to
    generate captions for ``novel scene compositions" successfully.
    The model has seen ``cat" and ``couch" while training, but it has not seen an image that contains these two with each other \cite{Lu2018NBT}.}
    \label{fig7}
\end{figure*}

Results for the Karpathy's split on the COCO dataset are reported in Table \ref{table1}.
Instances of images from the validation set of Karpathy's split
and generated captions for them are shown in Fig. \ref{fig5}, Note that in Fig. \ref{fig5}, Fig. \ref{fig6}, and Fig. \ref{fig7}, we are only showing the labels for
bounding box detection in images for visualization purposes only. The labels indicate what the object detector
thinks about the objects that reside in particular regions of the image. In practice, the captioning models do not use the detection labels. The object detectors are only
used to provide us with bounding box coordinates, using another CNN backbone of our choice, we can extract the visual features and feed them to a shared embedded space.
\par
A close look at the results reported in Table \ref{table1} reveals that our model improves the CIDER \cite{MS_COCO} score by
0.98, which indicates that the model is learning the saliency of the objects that should be mentioned in the captions better than NBT. 
The improvements on BLEU4 \cite{BLEU_2002} score indicate that the our proposed model is capable of handling
long range dependencies between different words of the generated captions better than NBT.
On the other hand, the improvement on BLEU1 \cite{BLEU_2002} score reveals that our model is also performing better in word prediction in general.
The improvements on SPICE \cite{SPICE} metric indicate that our model is performing better than NBT
in describing semantic relationships between the objects in the generated caption. The METEOR metric indicates the quality of the translation task between the ground truth
caption and the generated caption for unseen images.
Having these metrics gives an overview of how well the models are performing under particular splits on COCO dataset.
\par
\par
\begin{table}[!ht]\centering
    \caption {Results on COCO and Karpathy's split \cite{Karpathy_2015_CVPR}.}
    \begin{tabular}{lccccc}
        & & & \textbf{Metrics} \\ 
        \cline{2-6} 
        Model & BLEU1 & BLEU4 & CIDER & METEOR & SPICE \\ \hline
        NBT  & 73.84 & 32.64 & 100.71 & 25.79 & 18.92\\       
        NTT   & \textbf{73.93} & \textbf{32.92} & \textbf{101.69} & \textbf{25.8} & \textbf{18.99}\\
    \end{tabular}
    \label{table1}
\end{table}
\par
The original Novel split \cite{Hendricks_2016_CVPR} introduced by Hendricks et al. excludes all the
captions which contain the name of some particular objects. These names originally were chosen to be ``bottle'', ``bus'',
``couch'', “microwave”, “pizza”, “racket”, “suitcase” and “zebra”.
We report the results of our experiments on the Novel split for in-domain images.
Table \ref{table2} shows the results for this split. Fig. \ref{fig6},
shows examples of images from the validation set of the Novel split and generated captions for these images.
In this split, half of the original COCO validation set images are used for
validation randomly, and the rest is used for training. 
\par
\par
\begin{table}[!ht]\centering
    \caption {Results on COCO and Novel split \cite{Hendricks_2016_CVPR}.} 
    \begin{tabular}{lccccc}
        & & \textbf{Metrics} \\ 
        \cline{2-4}  
        Model & BLEU4 & CIDER & SPICE \\ \hline
        NBT & 30.79 & 93.83 & 18.17\\       
        NTT & \textbf{30.82} & \textbf{94.01} & \textbf{18.26}\\
    \end{tabular}
    \label{table2}
\end{table}
\par
The Robust split was created to evaluate generated captions for
novel scene compositions \cite{Lu2018NBT}. 
This split has 110,234 and 3,915 and 9,138 images in train,
validation and test portions of this split. The results of our experiments on Robust splits are shown in Table \ref{table3}.
Instances of generated captions for the images from the validation set of the Robust split are shown in Fig. \ref{fig7}.
\par
\par
\begin{table}[!ht]\centering
    \caption {Results on COCO and Robust split \cite{Lu2018NBT}.
    } 
    \begin{tabular}{lccc}
        & & \textbf{Metrics} \\ 
        \cline{2-4} 
        Model & BLEU1 & BLEU4 & CIDER \\ \hline
        NBT  & 73.28 & 31.2 & 92.1 \\       
        NTT   & \textbf{73.37} & \textbf{31.26} & \textbf{92.21} \\
    \end{tabular}
    \label{table3}
\end{table}
\par
\par
The overall results over three different splits suggest that twin cascaded attention
model in NTT improves the previously implemented attention
model in NBT, especially in larger domains. In other words, by looking at the results for Karpathy's \cite{Karpathy_2017}, Robust \cite{Lu2018NBT} and Novel \cite{Hendricks_2016_CVPR} splits,
we observe that the highest amount of improvement on CIDER score is achieved under Karpathy's split that has a larger amount of training data available. 
This could indicate over-fitting under the other two splits. We suspect that this over-fitting
is caused by the differences in the numbers of training examples under different splits.
The results of our experiments suggest that if we
employ more cascaded attention channels in
deep networks we could achieve better performance,
specifically in domains that include more training data.
\par
We improved the results in all metrics for the Karpathy's split \cite{Karpathy_2015_CVPR} on COCO.
We also improved the results in three metrics out of five for
Robust and Novel splits.
The results of the experiments clearly indicate that twin cascaded
attention model could further improve deep networks that employ attention
mechanisms with a single channel of attention in domains with sufficient amount of data.
\par
Our proposed method benefits from employing additional attention channels.
The results of our experiments suggest that in the near future, with larger amounts of GPU memory,
we could employ more attention channels and train such models in bigger
domains with larger amount of data to achieve better performance.
The results also suggest that a model with a single attention channel
could perform better in smaller domains with less amount of data.
Therefore, our proposed method is suitable when a particular
deep learning model is going to be used in a larger domain.
\par
Our proposed attention model could be considered a flexible structure that could be expanded up to
the current GPU memory limits. The key to successfully expanding
the proposed cascaded attention model with more attention channels lies in finding the proper dropout rates for the outputs of language LSTMs
and the joint LSTMs in the expanded model. We found that increasing the dropout rate for the language LSTMs and the joint LSTM creates
a decrementing refinement effect in the proposed cascaded attention model.
\par
\subsection{Alternative Calculations}\label{sec4.2}
\par
As we explained in Section \ref{3.4}, we modified the hypothesis and context gathering and distribution in Neural Twins Talk in order to
make the model suitable for Up-Down Attention model. The reason is that in our experiments we found that with
the original hypothesis and context vectors calculation methods, Neural Twins Talk would achieve slightly lower results under all metrics when applied on Up-Down Attention model.
Therefore, by modifying the way the hypothesis and context vectors are calculated, we successfully achieve better performance
for Neural Twins Talk applied on Up-Down Attention in comparison with the original Up-Down Attention model that utilizes
one channel of attention instead of two channels of attention in Neural Twins Talk.
\par
In this section, first we discuss the results for the NTT utilized over Up-Down attention with alternative calculations,
which follows what was explained in Eq. \ref{eq18} - Eq. \ref{eq20}, and is denoted as NTTUD-v1 in Table \ref{table4}.
Following that we discuss the results for NTT utilized over Up-Down Attention
with alternative calculations and having a third parallel attention channel without the joint LSTM following what was explained in Eq. \ref{eq21} and Eq. \ref{eq22}, which is denoted as NTTUD-v2 in Table \ref{table4}.
\par
We trained the original Up-Down Attention model \cite{Anderson2017up-down} and NTTUD-v1 and NTTUD-v2 models all from scratch without any pre-training.
The CNN backbone used for all three models explained in Section \ref{3.4} was pre-trained on ImageNet \cite{imagenet_cvpr09} without fine-tuning.
We trained all three models for 30 epochs, and the learning rate and other training details are identical to those used in experiments for NTT over NBT.
\par
\begin{table}[!ht]\centering
    \caption {Results on COCO and Karpathy's split \cite{Karpathy_2015_CVPR}.}
    \begin{tabular}{lccccc}
        & & & \textbf{Metrics} \\ 
        \cline{2-6} 
        Model & BLEU1 & BLEU4 & CIDER & METEOR & SPICE \\ \hline
        NTTUD-v1   & \textbf{76.51} & \textbf{34.49} & \textbf{111.24} & \textbf{27.49} & \textbf{20.62}\\
        NTTUD-v2  & 76.13 & 34.01 & \textbf{110.51} & \textbf{27.28} & \textbf{20.39}\\       
        Up-Down \cite{Anderson2017up-down} & 76.19 & 34.22 & 110.27 & 27.12 & 20.23\\           
    \end{tabular}
    \label{table4}
\end{table}
\par
The results of experiments in Table \ref{table4} clearly indicate that NTTUD-v1, which employs NTT over Up-Down Attention employing alternative calculations
performs better than NTTUD-v2, which uses a third parallel attention channel without a joint LSTM.
The performance improvement in NTTUD-v1 in comparison with NTTUD-v2 shows that having a joint LSTM is necessary for the Neural Twins Talk architecture.
At the same time we could observe that adding a third parallel channel without a joint LSTM improves the performance of a similar model with one channel of attention,
yet if the model is equipped with a joint LSTM, even with two parallel attention channels performs better.
We can also observe that both NTTUD-v1 and NTTUD-v2 are performing better than a similar model with one channel of attention.
This shows that adding parallel attention channels improves the performance of a deep learning model with attention and LSTM units.
This is what we also observed in experiment results for NTT over NBT.
\par
Another point worthy of being mentioned here is that NTTUD-v1 performs better under all metrics including the BLEU scores in comparison with
the Up-Down Attention model, whereas NTTUD-v2 which employs
three parallel attention channels without a joint LSTM, performs better under CIDER, SPICE and METEOR, while performing slightly worse
under BLEU scores in comparison with Up-Down Attention model.
This shows that the joint LSTM is helping the NTT model with learning the sentence structure
and grammatical correctness of the image captions generated by the model.
\par
Alternative calculations were clearly shown to be useful in gaining performance boost when NTT over NBT was employed over Up-Down Attention. 
Instead of changing the architecture of NTT we only modified the way the hypothesis
and context vectors are gathered and distributed in NTT over Up-Down Attention (NTTUD-v1).
Because the alternative calculations showed their effectiveness in performance gain when NTTUD-v1 was used, we also developed NTTUD-v2 to
investigate the importance of having a joint LSTM in NTT.
\par
Alternative calculations of hypothesis and context vectors over the same model with LSTMs and attention leads to better performance. This shows that
for every architecture containing LSTMs and attention there could be a subset of alternative calculation methods for hypothesis and context vectors in the model
that could further improve the performance of that model.
\par
Considering the effectiveness of alternative calculation methods we could conclude that these methods could potentially create another search space for
Neural Architecture Search (NAS) methods where each candidate in the search space of architectures
could have a subset of alternative calculation method candidates.
If these NAS models perform the search not only in architecture search space but also in alternative calculation methods search space, these models could potentially
find a better setting for an architecture where it performs better than the same architecture with conventional calculation methods in deep learning models.
\par
  \section{Conclusions}\label{sec5}
\par
We introduced a new attention model, namely twin cascaded attention model
that employs cascaded adaptive gates
and shows the importance of having multiple attention channels rather than having one attention channel in a deep learning model.
Looking at the results, we observe that these improvements have led to
performance gain, and these results also promise that in the near future, having more GPU memory,
we could employ more parallel attention channels to achieve better results. The results also suggest that employing more attention channels
demands more training data. In other words, we need more data to train more attention channels. Our proposed method promises
advancements and improvements in deep neural networks employing attention mechanisms
for image captioning and other similar tasks in vision-language.
\par

  \printbibliography

  \end{document}